\pdfoutput=1

\documentclass[11pt]{article}

\usepackage{acl}

\usepackage{times}
\usepackage{latexsym}

\usepackage[T1]{fontenc}

\usepackage[utf8]{inputenc}

\usepackage{microtype}

\usepackage{graphicx}

\title{Syntax-guided Neural Module Distillation to Probe\\ Compositionality in Sentence Embeddings}

\author{Rohan Pandey \\
  Language Technologies Institute \\
  Carnegie Mellon University \\
  \texttt{rspandey@cs.cmu.edu} \\}

\begin{document}
\maketitle
\begin{abstract}
Past work probing compositionality in sentence embedding models faces issues determining the causal impact of implicit syntax representations. Given a sentence, we construct a neural module net based on its syntax parse and train it end-to-end to approximate the sentence's embedding generated by a transformer model. The distillability of a transformer to a Syntactic NeurAl Module Net (SynNaMoN) then captures whether syntax is a strong causal model of its compositional ability. Furthermore, we address questions about the geometry of semantic composition by specifying individual SynNaMoN modules' internal architecture \& linearity. We find differences in the distillability of various sentence embedding models that broadly correlate with their performance, but observe that distillability doesn't considerably vary by model size. We also present preliminary evidence that much syntax-guided composition in sentence embedding models is linear, and that non-linearities may serve primarily to handle non-compositional phrases.
\end{abstract}

\section{Introduction}

The principle of semantic compositionality suggests that the meaning of a sentence should derive from its subconstituents in a regular, structured fashion \cite{montague1970english}. In recent years, transformers \cite{vaswani17} have become effective at producing sentential meaning representations useful for downstream tasks such as Natural Language Inference, Image-Text Matching, and Document Classification \cite{conneau17, radford21}. However, it has famously been conjectured that "You can't cram the meaning of a whole \%\&!\$\# sentence into a single \$\&!\#* vector" \cite{mooney14}. Since recent models do appear to capture sentence meaning effectively, one wonders how they compose arbitrarily many word meanings together such that their relational structure is captured in a single, fixed-dimensional sentence embedding.

\begin{figure}
    \centering
    \includegraphics[width=3in]{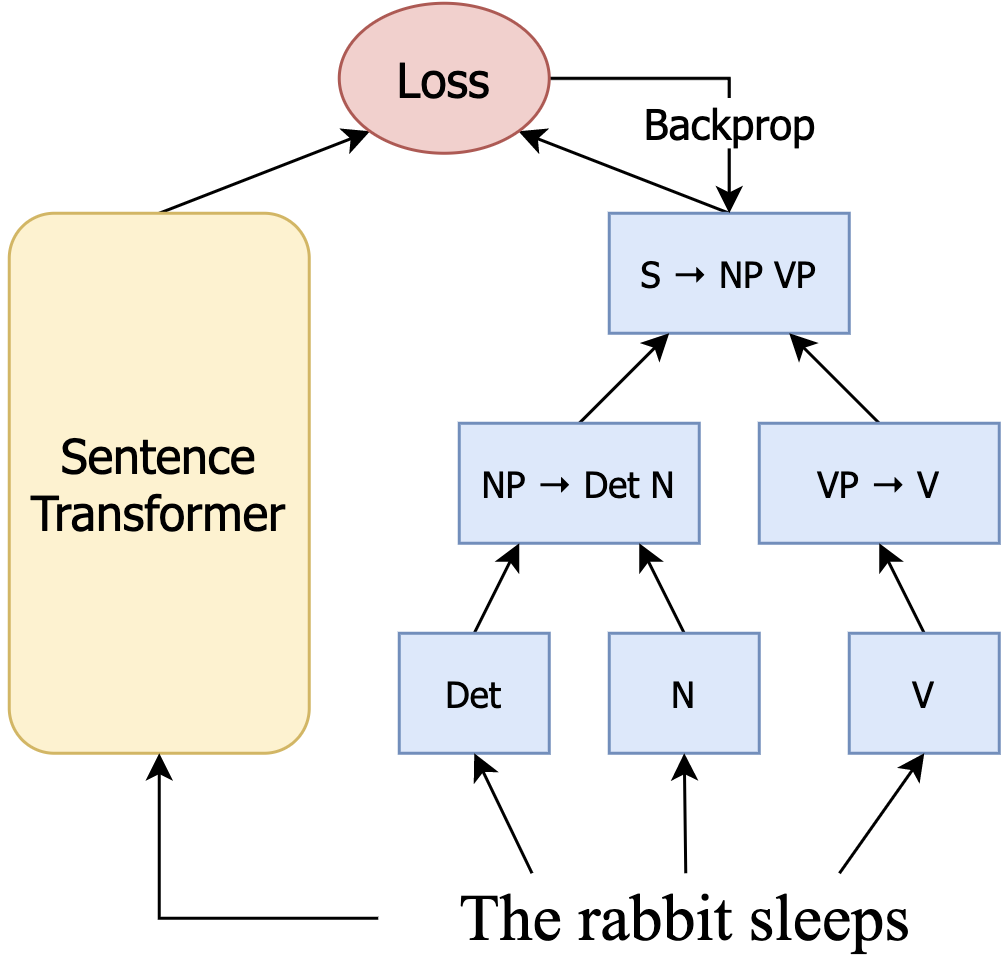}
    \caption{Distilling a transformer to a neural module net structured by the sentence's syntax}
    \label{fig:distillation}
\end{figure}


Much work has sought to probe these models for syntax representations and their causal relevance to embedding output. \citet{conneau18} train linear probes to determine if models encode syntactic features like tree distance and depth \cite{krasnowska19, hewitt19}. One line seeks out direct mappings between neural representations and tree structures \cite{mccoy18, chrupala19, jawahar-etal-2019-bert, soulos20, murty2023characterizing}. Other work raises methodological issues with probing \cite{eger19, zhu-rudzicz-2020-information} such as choice of formalism \cite{kuznetsov20} and semantic entanglement \cite{maudslay21}. \citet{ravichander21} raise the possibility that probing may identify causally un-used features; \citet{tucker21} partly address this concern to show that some syntactic features are causally relevant. Another line of work explores the geometry of semantic representations \cite{reif19, hernandez21} and the linearity \cite{baranvcikova19} of syntactic analogies \cite{zhu20}.

Rather than directly analyze sentence embedding models, Neural Module Nets \cite{andreas16} seek to improve compositionality by modularizing semantic functions. We see this effort as ultimately similar to probing for structural representations since the former explores whether explicit structure improves performance and the latter explores whether performant models implicitly learn structure. \citet{geiger21} discovers logical tree causal structures in BERT and \citet{wu21} then guides model distillation using this structure.

Our work builds on these findings by strictly taking syntax as the causal structure of sentential semantics and linearity as the geometry of syntax-guided composition; we conduct experiments to test the distillability of transformer-based sentence embedding models to a Syntactic NeurAl Module Net (SynNaMoN), an architecture we introduce that implements these two priors. The extent to which a model can be distilled to a SynNaMoN tells us about its internal syntax representations \& compositional ability.

\section{Methods}

\subsection{Syntactic Neural Module Net}
\begin{figure}[h]
    \centering
    \includegraphics[width=3in]{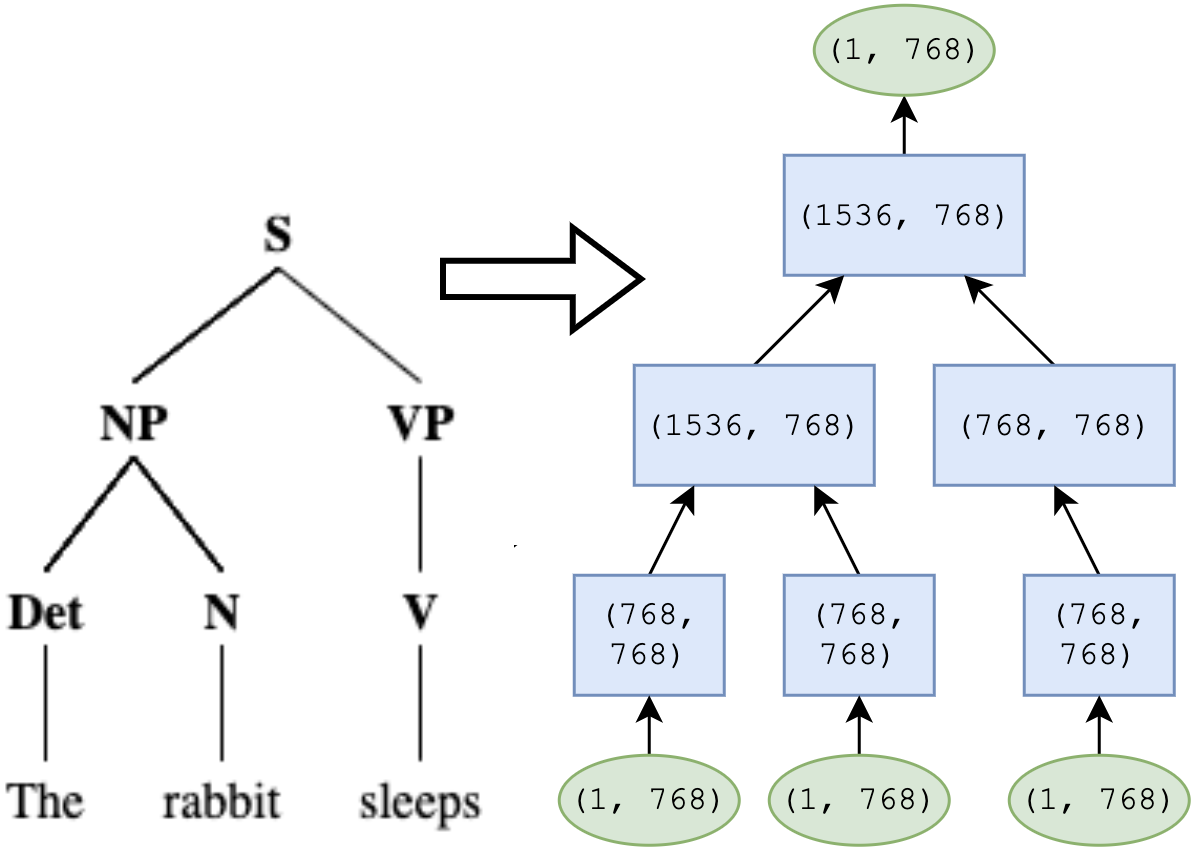}
    \caption{Constructing a sentence's SynNaMoN from its syntax tree; module input and output dimensionalities labeled on the right.}
    \label{fig:syn2mod}
\end{figure}

Unlike prior work \cite{andreas16qa, cirik18}, SynNaMoN modules don't approximate high-level objectives like `Find' or `Count' but rather correspond to specific syntactic rules like `S $\rightarrow$ NP VP' and `NP $\rightarrow$ DT JJ NN'. Each module receives an input of dimensionality $(1, N*D)$ where $N$ is the number of constituents on the syntax rule's right-hand side, and D is the dimensionality of the embedding space (768)—in other words, the input embeddings are concatenated. Though computationally more expensive, concatenation enables the module to learn an arbitrary function over the inputs rather than restricting it to a function over their sum or mean; this enables the module to converge on its ideal composition function which is likely not invariant under summation or averaging. Finally, though our implementation of SynNaMon includes `part-of-speech' modules at the bottom of the parse tree, one could conceivably remove this bottom layer with the hypothesis that the word embeddings already capture part-of-speech information.


\subsection{Internal Module Architecture}
\label{sec:internalarch}
To explore the geometry of semantic composition under syntax, we implement 3 module architectures: a linear layer (\textbf{Linear}), a linear layer + a ReLU activation (\textbf{Nonlin}), and a linear layer + ReLU + another linear layer (\textbf{Double}). We explore these 3 architectures to see whether syntax is enough of an inductive bias to linearly approximate sentence embeddings, or if adding non-linearities and additional layers considerably improves performance. The extent to which adding parameters improves our approximation of the teacher model beyond the syntactic structure alone could reveal how much isn't captured by this inductive bias.

\subsection{Linguistic Formalism}
We choose to use the Transformational Grammar presented by Penn Treebank \cite{marcus94}, but in principle any Constituency Grammar could be easily used with SynNaMoN, and Dependency Grammars can be adapted with some effort. Since prior work has shown how the choice of linguistic formalism can significantly influence probing results \cite{kuznetsov20}, we float the possibility of such an effect being at play in this work as well. If a student SynNaMoN fails to capture much of the teacher embedding model, perhaps it isn't because of the teacher's non-compositional causal structure, but rather because the formalism used to structure the SynNaMoN is inadequate. Indeed, recent state-of-the-art neural approaches to syntax parsing have learned grammatical tagsets that often differ starkly from human-produced syntactic theories \cite{kitaev2022learned}. We leave these problems to future work which may explore the exciting possibility that certain linguistic formalisms (perhaps even semantic rather than syntactic) are better proxies for a model's compositional structure than others.

\section{Experiments}
\label{sec:experiments}
Our main experiment runs 5 sentence embedding models (BERT-base \cite{devlin-etal-2019-bert}, MP-Net \cite{song2020mpnet}, GTR-T5-base, GTR-T5-large, and GTR-T5-xl \cite{ni2021gtr}) on 3 SynNaMoNs with differing internal architectures (Linear, Nonlin, Double; see Sec. \ref{sec:internalarch}). For BERT-base, we extract input word embeddings for each token and use the CLS token as the sentence embedding as is common practice. For the other 4 models, we encode each token alone to serve as its embedding and use the output as the sentence embedding. When words are encoded as more than 1 token, we compute the mean across the subtokens to serve as its word embedding.

In order to heuristically select a learning rate, 5 training runs were conducted with SynNaMoNs optimizing for BERT-base, and learning rate manually set at increments between $10^{-5}$ and $10^{-3}$. We finally chose a rate of $5\times10^{-5}$, but recognize from results that optimal learning rate will likely vary by teacher model \& SynNaMoN internal architecture. 
Analysis would best be reported on the optimal scores achieved by a SynNaMoN after hyperparameter tuning, but due to compute restrictions (1 NVIDIA K80 GPU with 12GB of RAM), this was unfeasible.

Additionally, due to the number of modules (originally ~900, each with ~1M parameters on average), we encountered frequent out-of-memory errors both on CPU \& GPU. Since each module corresponds to a syntax rule and is initialized upon encountering the rule in the dataset, we constrained our data to minimize the number of modules needed.

Specifically, we first constrained our trees to those of height 4 \& 5 (n=16492) in PTB, and then further constrained the trees to those that use a subset of the 300 most common production rules among them. This resulted in 1494 trees, from which we generated a train-validation split of 1250-244. Furthermore, we ensured that all the productions present in trees of the validation split were also included amongst trees in the training split. All this finally resulted in 273 production rules present in our dataset, and the instantiation of 273 modules.

\section{Results}
In Tab. \ref{tab:val_loss}, we present scores for all 5 sentence embedding models across the 3 SynNaMon architectures. We compute the average MSE between sentence embeddings in the complete dataset for each model and divide each model's MSE loss by this mean distance to normalize results. The normalized scores we present may intuitively be seen as the portion of variance in a model's sentence embeddings that a SynNaMoN fails to explain. From a probing perspective, the lower a model's score, the more it can be causally approximated by composition along syntactic lines.


\begin{table}[h]
    \centering
    \begin{tabular}{c|c|c|c}
         Sent. Emb. Model & Linear & Nonlin & Double  \\
         \hline
         BERT-base-CLS & .765 & 4.17 & .625 \\
         MP-Net-base & .606 & .963 & .538 \\
         GTR-T5-base & .541 & .844 & .499 \\
         GTR-T5-large & .550 & .898 & .502 \\
         GTR-T5-xl & \textbf{.536} & \textbf{.775} & \textbf{.498} \\
    \end{tabular}
    \caption{Best validation MSE loss of sentence embedding models on each SynNaMoN probe, normalized by chance-level MSE between embeddings}
    \label{tab:val_loss}
\end{table}

First, notice that GTR-T5-xl outperforms all the other models across all the SynNaMoN architectures. This seems to confirm our expectation that larger models should produce more compositional sentence embeddings. However, GTR-T5-xl only marginally outperforms other sizes of GTR-T5 (except on Nonlin, for which it does far better), suggesting that size actually isn't a significant factor in compositionality. The lower performance of GTR-T5-large further corroborates this, but considering its anomalously lower average embedding MSE, the issue requires more work. The fact that GTR-T5 models all display high compositionality despite variance in size suggests something about their architecture or training approach is important—perhaps the representational bottleneck.

All 3 GTR-T5 models perform better than MP-Net, which in turn outperforms BERT CLS. This first fact is slightly surprising considering that on standard sentence representation tasks \cite{reimers19}, MP-Net (63.30) marginally outperforms all GTR-T5 (base: 59.40, large: 62.38, xl: 62.88) models. Evaluation of these sentence embedding models on large-scale, human-interpretable compositionality tasks may reveal that GTR-T5 does indeed produce better compositional representations than MP-Net. Although BERT's CLS token embedding is widely used for sentence representation, these results show that it fails to capture nearly as much compositional information as more targeted sentence embedding models.

Next, observe that although distilling to a Double SynNaMoN is intuitively easier than to a Linear SynNaMoN due to increased parameterization, there aren't always major improvements in distillability. It is possible that the geometric expressivity of the Double SynNaMoN will kick in with scaling of training data, but we hypothesize that this Double score will still approach a limit for all sentence embedding models. This is because syntax only describes a subset of sentence meaning, and the strictness of SynNaMoN's structure prevents this non-compositional component from being learned.

For example, a strictly syntactic compositional interpretation of "village on the river", would represent the village as being literally on top of the river since this is the semantic geometry learned for syntactic structures of the form "NP on NP". A SynNaMoN that includes non-linearities may better learn the geometry of this non-literal "on" relation, but a transformer model would best learn to handle non-compositional phrases due to its lack of strict syntactic constraints. Our broader takeaway from comparing Linear \& Double scores is that much composition along syntactic lines is linear, and non-linearities in transformers primarily serve a purpose other than syntax-guided composition—perhaps in handling non-compositional phrases.


\begin{figure}
\includegraphics[width=3in]{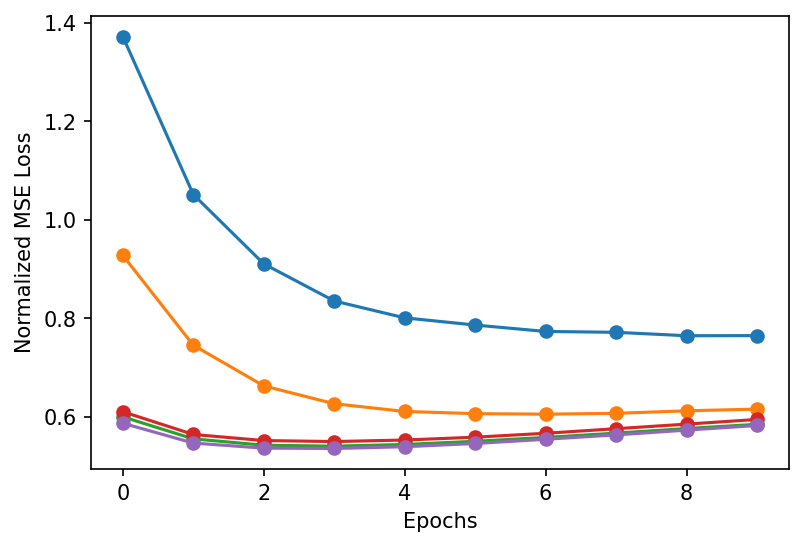}
\caption{Normalized validation learning curves for Linear SynNaMoN on sentence embedding models (blue: BERT, orange: MP-Net, red: GTR-T5-large, green: GTR-T5-base, purple: GTR-T5-xl)}
\label{fig:loss_curves}
\end{figure}

On a less theoretical note, we observe that our learning curves for Linear SynNaMoN on GTR-T5 (Fig. \ref{fig:loss_curves}) are clearly overfitted due to fixing hyperparameters as mentioned in Sec. \ref{sec:experiments}. We remediate this issue in Tab. \ref{tab:val_loss} by reporting the best scores (minimum across epochs) for each learning curve. Since we want to construct the best possible SynNaMoN for a transformer model (as this most accurately reveals the transformer's distillable compositional ability), scores could be slightly improved with further hyperparameter tuning.

\subsection{Analysis}

Finally, we explore a single module to determine whether its compositional geometry meets intuitive notions of semantic generalization. Due to methodological difficulties with assessing a single module extracted from our end-to-end training paradigm, we train a Linear module for `NP $\rightarrow$ Det N' on its own. Determiner-noun composition intuitively lies on a spectrum with adjective-noun composition on the other end and quantifier-noun composition in between. While quantifiers like `some' and `all' seem more like determiners, other quantifiers like `several' and `twelve' appear more comparable to adjectives like `swarming' and `grouped'. Intuitively then, we should expect the geometry of quantifier-noun composition to be intermediate to determiners and adjectives.

\begin{figure}
    \centering
    \includegraphics[width=3in]{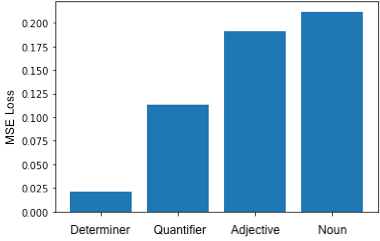}
    \caption{Generalization ability of Determiner Phrase module's linear geometry varies by part-of-speech}
    \label{fig:determiners}
\end{figure}

And this behavior is precisely what we find in our `NP $\rightarrow$ Det N' module. Since it's trained on determiners, it obviously has the lowest MSE for this part-of-speech; we include noun-noun pairs (e.g. `tree cow') as a control. As seen in Fig. \ref{fig:determiners}, the module generalizes to quantifiers intermediately to determiners and adjectives. This demonstrates how SynNaMoN modules may enable interesting analyses of the compositional geometry of syntactic operations in sentence embedding models.

\section{Conclusion}
The human ability to apprehend the unitary meaning of a sentence corresponds to a neural model's ability to construct compositional sentence embeddings. In this work, we introduced Syntactic Neural Module Nets and used it in a distillation approach to assess how well syntax explains the sentential semantics computed by a transformer model. We showed that some models are more compositional by this metric, syntax-guided composition is largely linear, and modules learn composition functions that correspond to our semantic intuition. 

Future work could explore this approach's alignment with other compositionality metrics and the non-compositional semantics left uncaptured by SynNaMoNs. We are also interested in how SynNaMoNs of different linguistic formalisms vary in distillability, as well as other potential use cases of SynNaMoNs beyond probing.

\section{Limitations \& Ethics Statement}

Since longer sentences have more complex syntax, they require more modules on GPU and can run into out-of-memory issues. However, there is a hard upper-bound on total number of modules since there are limited syntax rules in the grammar. In addition, we may never need to train on long sentences if all modules can be effectively trained on short sentences and then generalize compositionally.

As an approach to probing language models, SynNaMoN contributes to an ethical NLP vision that seeks to address how models learn human biases that have societal effects from corpus data. Understanding syntax representations in models could be important in such a pursuit since some of these bias effects are syntactically mediated. For example, LMs with gender role biases could internally represent these biases as syntactic gender agreement e.g. `man' agrees with 'doctor' and `woman' agrees with `nurse' \cite{prates2020assessing}. By understanding the causal structure of sentential semantics in LMs, we can better disentangle syntax from spurious correlations transmitted by societal structures.

\section{Acknowledgements}

I am grateful for the advice provided by Tom McCoy, Emmy Liu, Chris Potts, and Paul Smolensky on questions of compositionality \& probing. I also appreciate the intellectual company of Aryaman Arora \& Justus Mattern at the onset of this project, and to Uncle Ike for providing us with a truly inspirational set of trees.

\bibliographystyle{acl_natbib}
\bibliography{custom}

\end{document}